\documentclass[10pt,twocolumn,letterpaper]{article}

\usepackage{cvpr}
\usepackage{times}
\usepackage{epsfig}
\usepackage{graphicx}
\usepackage{amsmath}
\usepackage{amssymb}
\usepackage{color}
\usepackage{booktabs}
\usepackage{array}
\usepackage{lipsum}
\usepackage[breaklinks=true,bookmarks=false]{hyperref}

\cvprfinalcopy % *** Uncomment this line for the final submission

\def\cvprPaperID{1453} % *** Enter the CVPR Paper ID here

\author{\Large 
Xuangeng Chu\textsuperscript{\rm 1*},
Anlin Zheng\textsuperscript{\rm 2*},
Xiangyu Zhang\textsuperscript{\rm 2*\dag},
Jian Sun\textsuperscript{\rm 2}\\
\textsuperscript{\rm 1}Peking University\quad
\textsuperscript{\rm 2}MEGVII Technology\\
\texttt{\small xg\_chu@pku.edu.cn, \{zhenganlin, zhangxiangyu, sunjian\}@megvii.com}\\
}

\begin{document}

%%%%%%%%% TITLE
\title{Detection in Crowded Scenes: One Proposal, Multiple Predictions}
\maketitle

{\let\thefootnote\relax\footnotetext{* Equal contribution. The work is done when Xuangeng Chu was an intern in MEGVII Technology.}}
{\let\thefootnote\relax\footnotetext{\dag\ Corresponding author.}}
{\let\thefootnote\relax\footnotetext{This work is supported by The National Key Research and Development Program of China (2018YFC0831700) and Beijing Academy of Artificial Intelligence (BAAI). }}

%%%%%%%%% ABSTRACT
\begin{abstract}
   We propose a simple yet effective proposal-based object detector, aiming at detecting highly-overlapped instances in crowded scenes. The key of our approach is to let each proposal predict a set of correlated instances rather than a single one in previous proposal-based frameworks. Equipped with new techniques such as EMD Loss and Set NMS, our detector can effectively handle the difficulty of detecting highly overlapped objects. On a FPN-Res50 baseline, our detector can obtain 4.9\% AP gains on challenging CrowdHuman dataset and 1.0\% $\text{MR}^{-2}$ improvements on CityPersons dataset, without bells and whistles. Moreover, on less crowed datasets like COCO, our approach can still achieve moderate improvement, suggesting the proposed method is robust to crowdedness.
   Code and pre-trained models will be released at https://github.com/megvii-model/CrowdDetection.
\end{abstract}

%%%%%%%%% BODY TEXT
\section{Introduction}
\label{sec:intro}
Proposal-based framework has been widely used in modern object detection systems \cite{ren2015faster,girshick2015fast,girshick2014rich,lin2017feature,lin2017focal,liu2016ssd,he2014spatial,dai2016rfcn,dai2017deformable,cai2019cascadercnn,he2017mask,yolov3,sniper2018,snip_object}, both for one-stage~\cite{liu2016ssd,yolov3,lin2017focal,dssd} and two/multi-stage~\cite{ren2015faster,lin2017feature,dai2016rfcn,dai2017deformable,he2017mask,cai2019cascadercnn} methods. The paradigm in general has a two-step pipeline: first, generating \emph{over-complete} object proposals in handcraft (e.g. predefined \emph{anchors} \cite{yolov3,liu2016ssd,lin2017focal}) or learnable (e.g. \emph{RPNs} \cite{ren2015faster,lin2017feature,he2017mask}) manner; then, predicting a \emph{single} instance corresponding to each proposal box with a confidence score and a refined location. To remove duplicate predictions, methods such as \emph{Non-maximum Suppression (NMS)} are usually required for post-processing. 

\begin{figure}[!t]
\begin{center}
 \includegraphics[width=1.\linewidth]{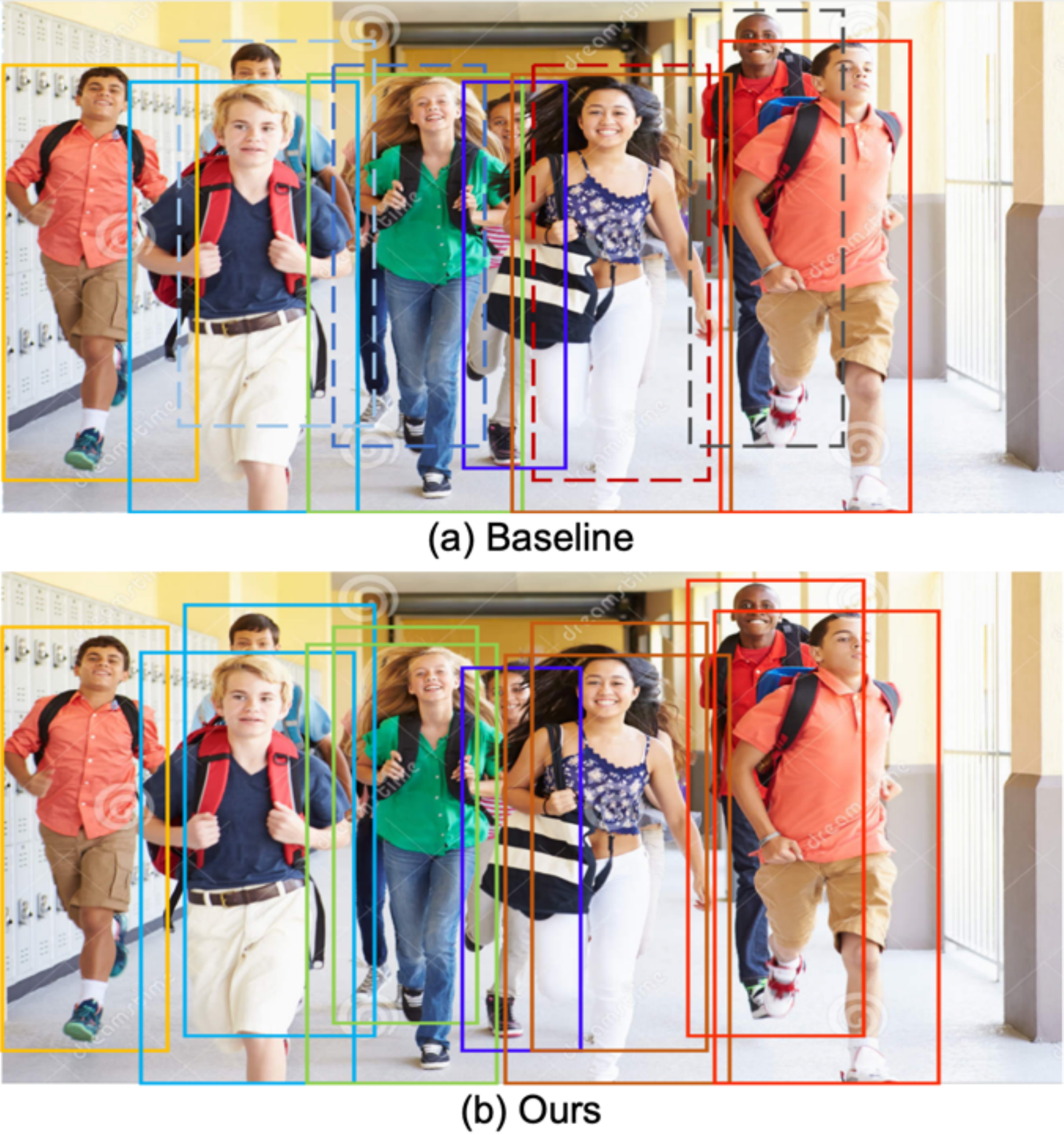}
\end{center}
   \caption{Human detection in crowds. (a) Results predicted by \emph{FPN} \cite{lin2017feature} baseline. The dashed boxes indicate the missed detections (suppressed by \emph{NMS} mistakenly). (b) Results of our method applied to FPN. All instances are correctly predicted. Boxes of the same color stem from the identical proposal (best viewed in color).   }
\label{fig:illustration}
\end{figure}

Although proposal-based approaches have achieved state-of-the-art performances \cite{he2017mask,cai2019cascadercnn,lin2017feature,lin2017focal} in popular datasets such as \emph{COCO} \cite{lin2014microsoft} and \emph{PASCAL VOC} \cite{pascalvoc}, it is still very challenging for crowded detection in practice. Fig.~\ref{fig:illustration} (a) shows a common failure case: the detector fails to predict instances heavily overlapped with others (indicated in dashed box). 

This kind of typical failure in crowded scenes is mainly ascribed to two reasons. First, highly overlapped instances (as well as their associated proposals) are likely to have very similar features. As a result, it is difficult for a detector to generate distinguishing prediction for each proposal respectively (illustration is show in Fig.~\ref{fig:method} for a concrete example). Second, since instances may heavily overlap each other, the predictions are very likely to be mistakenly suppressed by \emph{NMS}, as depicted in Fig.~\ref{fig:illustration} (a). 

Previous works have tried to address this issue from different perspectives, such as sophisticated NMS \cite{softnms,he2019bounding,adaptiveNMS,learningnms,sequentialcontext,hosang2016convnet}, new loss functions \cite{occludedattention,repulseloss}, re-scoring \cite{hu2017relation}, part-based detectors \cite{occludedattention,tian2015deep,Zhou_2018_ECCV,chi2019pedhunter}. However, as we will analyze later (Sec.~\ref{sec:limitations}), current works are either too complex or less effective for handing highly-overlapped cases, or degrading the performance of less-overlapped cases. 

In this paper, a new scheme is introduced to handle this difficulty: for each proposal box, instead of predicting a single instance, as usual, we suggest predicting \emph{a set of instances} that might be highly overlapped, as described in Fig.~\ref{fig:method}. With this scheme, the predictions of nearby proposals are expected to infer the \emph{same set} of instances, rather than \emph{distinguishing individuals}, which is much easy to be learned. We also introduce several techniques in the new scheme. Firstly, a \emph{EMD loss} is proposed to supervise the learning of instance set prediction. Secondly, a new post-processing method named \emph{Set NMS} is introduced to suppress the duplicates from different proposals, which aims at overcoming the drawbacks of na\" ive NMS in crowd scenes. Lastly, an optional \emph{refinement module} (RM) is designed to handle the potential false positives. 

Our method is simple and almost cost-free. It is applicable to all the proposal-based detectors such as \cite{ren2015faster,lin2017feature,lin2017focal,he2017mask}. The major modification is adding a prediction branch, which only brings negligible extra cost. But the improvement is significant: on \emph{CrowdHuman} \cite{shao2018crowdhuman} dataset, our method boosts the AP score by \textbf{4.5\%} (without refinement module) or \textbf{4.9\%} (with refinement module); in addition, the recall on crowded instances improves by \textbf{8.9\%}. More importantly, fewer false positive appears, suggested by the slightly improved $\text{MR}^{-2}$ index even without refinement module. Besides, on less crowded datasets, our method can still obtain moderate gains. For example, on \emph{CityPersons} we achieve 0.9\% and 1.0\% improvements in AP and $\text{MR}^{-2}$ over the baseline; and on \emph{COCO}, \cite{lin2014microsoft} it obtains 1.0\% higher AP score. All experiments conducted on different datasets illustrate that our method can handle all scenes gracefully, regardless of the crowdedness.

%-------------------------------------------------------------------------
\section{Background}
\label{sec:limitations}

\begin{figure}[t]
	\begin{center}
		\includegraphics[width=0.8\linewidth]{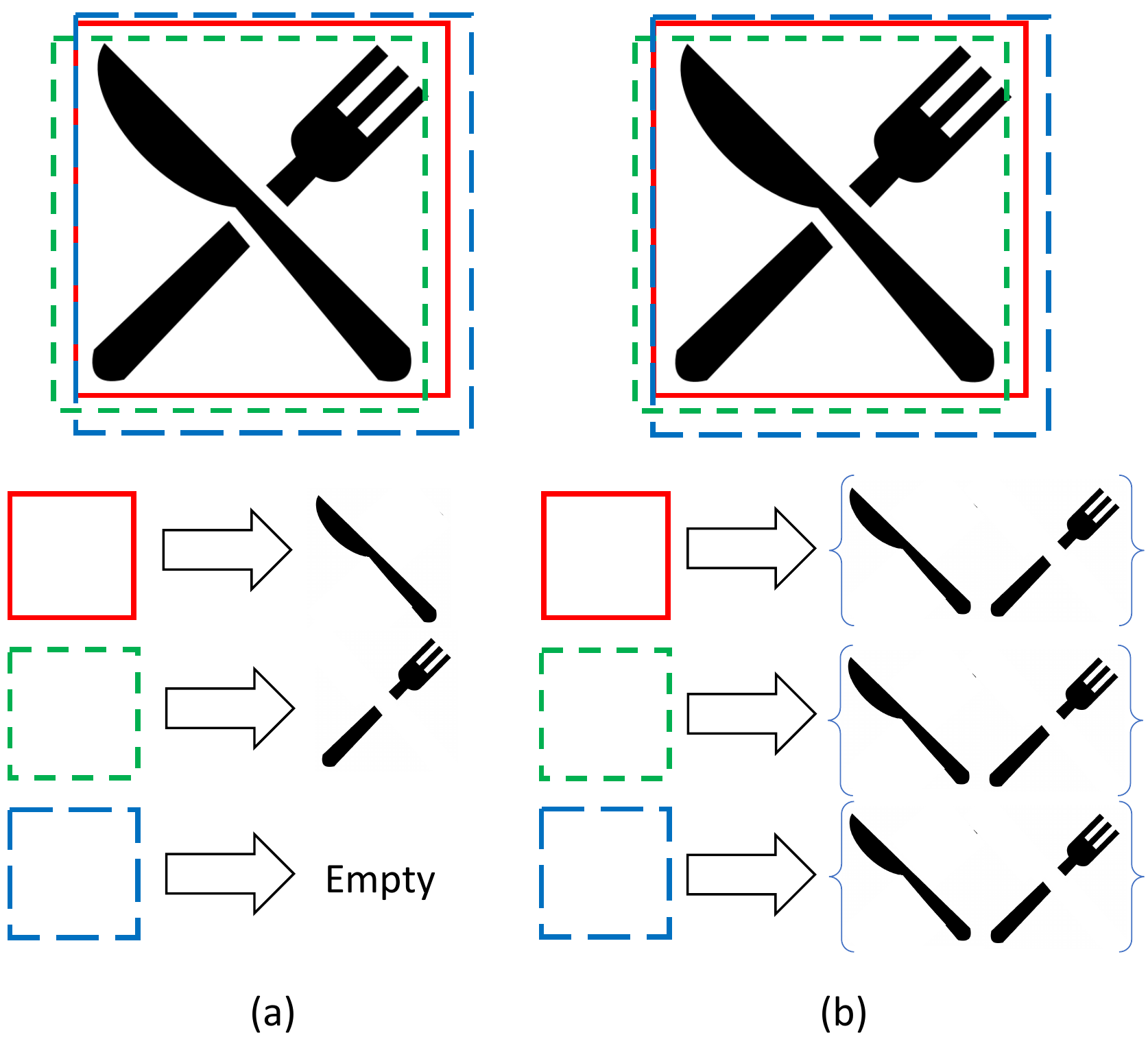}
	\end{center}
	\caption{A typical case in crowded detection. A knife and a fork share almost the same bounding boxes. Three proposal boxes (red, green and blue, best viewed in color) are heavily overlapped. (a) Single predication paradigm (see Sec.~\ref{sec:limitations}). Each proposal box is expected to predict a \emph{single instance} \cite{ren2015faster,lin2017feature,lin2017focal,liu2016ssd,yolov3,he2017mask} (may be empty) , which is intrinsically difficult as the proposals share very similar feature. Moreover, after NMS, it is very likely that only one prediction survives. (b) In our approach, each proposal predicts \emph{a set of instances}. Our \emph{Set NMS} can easily remove the duplicate prediction sets together (not illustrated in the figure).}
	\label{fig:method}
\end{figure}

As mentioned in the introduction, the paradigm of proposal-based object detectors mainly involves two steps: the first step is \emph{proposal box generation}, which could be achieved by \emph{Selective Search}~\cite{girshick2014rich,girshick2015fast}, predefined/learnable \emph{anchors} \cite{ren2015faster,yolov3,liu2016ssd,lin2017focal,yang2018metaanchor,wang2019region,zhong2019cascade} or \emph{Region Proposal Networks (RPNs)}~\cite{ren2015faster,lin2017feature,he2017mask,dai2016rfcn,cai2019cascadercnn}, etc. The second step is \emph{instance prediction}, i.e. predicting the refined detection results corresponding to each proposal box. We primarily focus on the second step in this paper.

For \emph{instance prediction}, current state-of-the-art object detection frameworks~\cite{ren2015faster,lin2017feature,lin2017focal,liu2016ssd,yolov3} usually attach a \emph{detection function} to each proposal box, which is used to determine whether the proposal is associated to some ground truth instance; if true, further to predict the corresponding class label and the refined bounding box for the object. This mechanism implies that each proposal box corresponds to \textbf{single} ground truth (usually the one with the largest overlap to the proposal box). Therefore, the proposal boxes have to be over-completed to ensure every instance has a chance to be detected, which introduces many duplicates to the predictions. As a result, duplicate removal methods such as \emph{Non-Maximum Suppression (NMS)} are necessary for those frameworks to filter out the duplicate results.

Although the above paradigm seems to obtain outstanding results on some benchmarks such as \emph{COCO} \cite{lin2014microsoft} and \emph{PASCAL VOC} \cite{pascalvoc}.  It suffers from missing detection in crowded scenarios due to post-processing methods, \eg NMS. Fig.~\ref{fig:illustration} (a) shows an example: people in the dashed boxes are suppressed by the nearby boxes mistakenly. Thus, several approaches or workarounds have been proposed to address this limitation, which can be categorized as follows:

\paragraph{Advanced NMS.}
The effectiveness of na\"ive NMS is based on the assumption that multiple instances rarely occur at the same location, which is no longer satisfied in the crowded scenes. Several improved NMS approaches have been proposed. For example, \emph{Soft-NMS} \cite{softnms} and \emph{Softer-NMS} \cite{he2019bounding} suggest decaying the confidence score of the neighboring predictions for suppression rather than directly discard them. \cite{optimizedpedestrian} employs Quadratic Binary Optimization to predict instances, taking advantage of the prior distribution of ground truth's sizes. However, such heuristic variants of NMS are not always valid under different circumstances.  Thus, more complex mechanisms may be introduced, for example, \cite{learningnms,sequentialcontext} uses a neural network for more sophisticated and data-dependent duplicate removal. Although these methods raise the upper bound of na\"i've NMS, the pipeline becomes much more complex and costly in computation. Other works such as \cite{adaptiveNMS,hosang2016convnet} propose to predict different NMS thresholds for different bounding boxes. As the major drawback, they need an extra structure for IoU/density estimation, which introduces more hyper-parameters. Besides, it is still difficult to distinguish heavily overlapped boxes as in Fig~\ref{fig:method} (a).

\paragraph{Loss functions for crowded detection. }
A few previous works propose new loss functions to address the problem of crowded detection. For example, \cite{zhang2018occlusion} proposes \emph{Aggregation Loss} to enforce proposals to be close and locate compactly to the corresponding ground truth. \cite{repulseloss} proposes \emph{Repulsion Loss}, which introduces extra penalty to proposals intertwine with multiple ground truths. The quality of detections in crowded scenes is improved with the help of these loss functions. However, since traditional NMS is still needed in the frameworks, it is still difficult to recall the overlapped instances illustrated in Fig~\ref{fig:method} (a).

\paragraph{Re-scoring. }
In many detection frameworks \cite{ren2015faster,liu2016ssd,lin2017feature,lin2017focal} a proposal box is bound to a ground truth as long as the overlap is larger than a given threshold, which usually leads to a \emph{many-to-one} relation between proposals and ground-truth instances thus NMS is required to remove the duplicate proposals. Instead, if we redesign the loss function to encourage \emph{one-to-one} relation, the NMS procedure may be eliminated to avoid miss detection. We name the method \emph{re-scoring}. Some of the previous works follow the idea. For example, in \cite{yolo,redmon2017yolo9000}, each ground-truth instance is associated strictly to one proposal box during training. However, in the architectures of \cite{yolo,redmon2017yolo9000}, due to lack of connections between proposals, the prediction may be ambiguous, because it is not sure for one proposal to determine whether the related instance has been predicted by another proposal. Actually, in \cite{yolo,redmon2017yolo9000} NMS is still involved. 
\emph{RelationNet} \cite{hu2017relation}, instead, explicitly models the relations between proposals, which is supposed to overcome the limitations of \cite{yolo,redmon2017yolo9000}. Using re-scoring, RelationNet obtains outstanding performance on \emph{COCO}~\cite{lin2014microsoft} dataset even without NMS. However, in a more crowded dataset \emph{CrowdHuman} \cite{shao2018crowdhuman}, we find RelationNet with re-scoring performs relatively poor (see Sec.~\ref{sec:exp} for details). It may be because on CrowdHuman dataset, proposals have to be much denser than those on COCO. As a result, the re-scoring network needs to generate different predictions from very close proposals (so their features and relations are also very similar, as shown in Fig.~\ref{fig:method} (a)), which is infeasible for neural networks. 

There are other approaches on crowded detection, for example, \emph{part-based detectors} \cite{occludedattention,tian2015deep,Zhou_2018_ECCV,chi2019pedhunter}, which is mainly used in detecting special instances such as pedestrian. The discussions are omitted in this paper. 

In conclusion, based on the above analyses, we argue that object detection in the crowded scene may be \emph{fundamentally} difficult, or at least \emph{nontrivial} and complex for the mentioned existing proposal-based frameworks. The key issue lies in the basic paradigm of predicting only one instance for each proposal box. It inspires us to explore new \emph{instance prediction} schemes, i.e. \textbf{multiple instance prediction} for each proposal. 

\section{Our Approach: Multiple Instance Prediction}
\label{sec:ourapproach}
Our approach is motivated by the observation: consider there are multiple objects heavily overlapped with each other, like the case in Fig.~\ref{fig:method}; if one proposal corresponds to any of the objects, it is very likely to overlap all the other objects. So, for such a proposal box, rather than predict a single object, why not predict them \emph{all}? Formally, for each proposal box $b_i$, the new scheme suggests predicting the correlated \textbf{set} of ground-truth instances $\mathrm{G}(b_i)$ instead of a single object:
\begin{equation}
\mathrm{G}(b_i) = \left\{ g_j \in \mathcal{G} | \mathrm{IoU} (b_i, g_j)\ge \theta \right\}, 
\label{equ:gt_set}
\end{equation}
where $\mathcal{G}$ is the set of all the ground truth boxes and $\theta$ is a given threshold of \emph{intersection-over-union} (IoU) ratio. Fig.~\ref{fig:method} (b) visualizes the concept. Compared with the previous \emph{single-instance-prediction} framework, we find our new scheme may greatly ease the learning in crowded scenes. As shown in Fig.~\ref{fig:method} (b), all of the three proposal boxes are assigned to the same set of ground-truth instance -- it is a feasible behavior since the three proposals actually share almost the same feature. While for the previous single-instance-prediction paradigm (Fig.~\ref{fig:method} (a)), each proposal has to produce distinguishing predictions, which might be intrinsically difficult.  
 
\begin{figure}[t]
\begin{center}
 \includegraphics[width=0.9\linewidth]{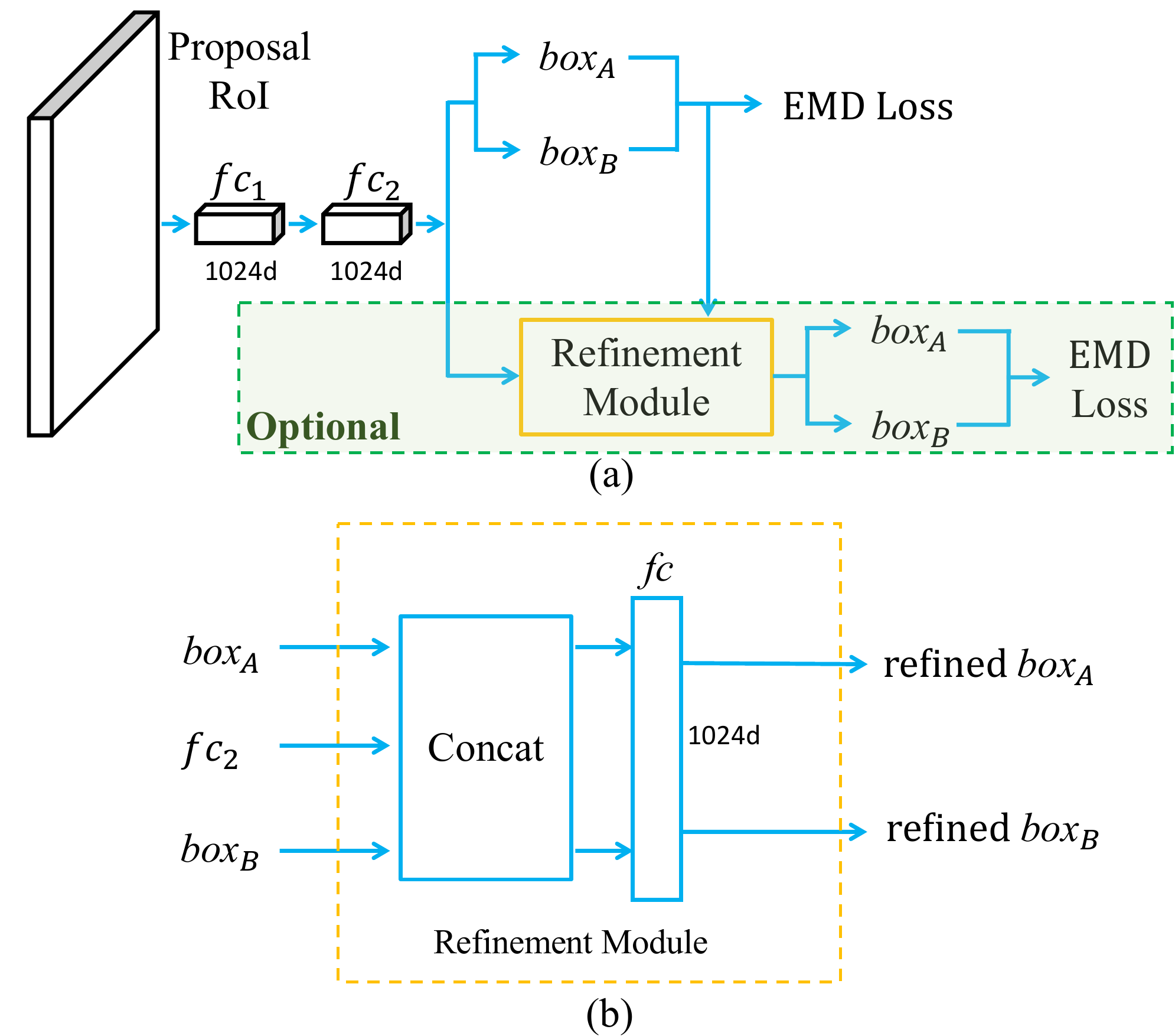}
%  \caption{}
\end{center}
   \caption{Overall architecture. (a) $box_A$ and $box_B$ are the two instances predicted by one proposal, using our EMD Loss. \emph{Refinement module} is an optional step. (b) The refinement module concatenates feature and box information to optimize the results.}
\label{fig:arch}
\end{figure}

We introduce the details of our approach as follows:

\paragraph{Instance set prediction. }
For each proposal box $b_i$, most of the modern proposal-based detection frameworks \cite{ren2015faster,lin2017feature,lin2017focal,liu2016ssd,he2017mask} employ a \emph{detection function} to predict a pair $(\mathbf{c}_i, \mathbf{l}_i)$ to represent the associated instance, where $\mathbf{c}_i$ is the class label with confidence and $\mathbf{l}_i$ is the relative coordinates. In our approach, to predict a set of instances, we introduce a simple extension -- just by introducing $K$ detection functions to generate a set of predictions $\mathrm{P}(b_i)$:
\begin{equation}
\mathrm{P}(b_i) = \left\{ (\mathbf{c}_i^{(1)}, \mathbf{l}_i^{(1)}),  (\mathbf{c}_i^{(2)}, \mathbf{l}_i^{(2)}), \dots ,  (\mathbf{c}_i^{(K)}, \mathbf{l}_i^{(K)}) \right\}, 
\label{equ:pred}
\end{equation}
where $K$ is a given constant standing for the maximum cardinality of $\mathrm{G}(b_i)$ in the dataset (see Eq.~\ref{equ:gt_set}). $\mathrm{P}(b_i)$ can be simply implemented by introducing extra prediction branches in most of the existing detection frameworks \cite{ren2015faster,lin2017feature,lin2017focal,he2017mask,liu2016ssd}, which is shown in Fig.~\ref{fig:arch} (a). Note that even though $K$ is fixed for all proposals, the network could still predict some $\mathbf{c}_i^{(k)}$ to background class, representing that the $k$-th detection function does not predict instance for the proposal $b_i$.

\paragraph{EMD loss. }
We aim to design a loss $\mathcal{L}(b_i)$ to minimize the gap between predictions $\mathrm{P}(b_i)$ and ground-truth instances $\mathrm{G}(b_i)$ corresponding to the proposal $b_i$, which can be cataloged into the problem of \emph{set distance measurement}. Similar problems have been discussed in some early object detection papers, such as \cite{szegedy2014scalable,erhan2014scalable,stewart2016end}. Inspired by them, we design the following \emph{EMD loss} to minimize the \emph{Earth Mover's Distance} between the two sets:
\begin{equation}
	\mathcal{L}(b_i) = \min_{\pi \in \Pi} \sum_{k=1}^{K} \left[ 
	\mathcal{L}_{cls} (\mathbf{c}_i^{(k)}, g_{\pi_k}) + \mathcal{L}_{reg} (\mathbf{l}_i^{(k)}, g_{\pi_k})
	\right]
\label{equ:loss}
\end{equation}
where $\pi$ represents a certain permutation of $(1, 2, \dots, K)$ whose $k$-th item is $\pi_k$; $g_{\pi_k} \in \mathrm{G}(b_i)$ is the $\pi_k$-th ground-truth box; $\mathcal{L}_{cls}(\cdot)$ and $\mathcal{L}_{reg}(\cdot)$ are classification loss and box regression loss respectively, following the common definitions as \cite{ren2015faster,liu2016ssd,lin2017feature,lin2017focal,he2017mask}. Note that in Eq.~\ref{equ:loss}, we assume $|\mathrm{G}(b_i)|=K$; if not, we add some ``dummy'' boxes (whose class label is regarded as background and without regression loss) to $\mathrm{G}(b_i)$ until it is satisfied. Intuitively, the formulation in Eq.~\ref{equ:loss} implies to explore all possible one-to-one matches between predictions and ground truths, thus finding the ``best match'' with the smallest loss. It is also worth noting that if $K=1$, Eq.~\ref{equ:loss} becomes equivalent to the loss in traditional \emph{single-instance-prediction} frameworks, implying that our EMD loss is a natural generalization to the commonly-used detection loss \cite{ren2015faster,liu2016ssd,lin2017feature,he2017mask}. 

\paragraph{Set NMS. }
In our approach, although each proposal is able to predict multiple associated instances, if na\"ive NMS is still involved for post-processing it is impossible to detect objects effectively in crowded scenes. Fortunately, because of the \emph{EMD loss}, instances predicted by one proposal are expected to be \emph{unique} by definition. In other words, duplicates exist only between the predictions from different proposals, as illustrated in Fig.~\ref{fig:method} (b). With this prior, we introduce a simple patch to na\"ive NMS pipeline, named \emph{Set NMS} -- each time before one box suppressing another one in the NMS algorithm, we insert an additional test to check whether the two box come from the same proposal; if yes, we skip the suppression. Experiments in Sec.~\ref{sec:exp} also suggest that only when \emph{multiple-instance-prediction} and \emph{Set NMS} are used together can our approach achieve significant improvement in crowded detection. 

\paragraph{Refinement module. }
In our approach, each proposal is expected to generate a set of instances rather than a single one, which may suffer from increase in false positives since more predictions are generated. %($\mathrm{G}(b_i)$, which is \emph{unordered}); however, a typical neural network can only directly output an \emph{ordered} sequence for $\mathrm{P}(b_i)$. Even though our \emph{EMD loss} in Eq.~\ref{equ:loss} has greatly solved the permutation problem, in the appendix, we have proved in mathematics that the mismatch may cause some false positive detections. 

Although the failure cases are rarely observed in our experiments on real images, we introduce an \emph{optional} refinement module in case of the risk, as shown in Fig.~\ref{fig:arch} (b). The module simply takes the predictions as input, combining them with proposal feature, then performs a second round of predicting. We expect the refinement module to correct the possible false predictions. 

\paragraph{Discussion: relation to previous methods.  }
Predicting multiple instance is not new. \emph{Double-person detector} \cite{Tang2014}  models person pairs in the \emph{DPM} \cite{felzenszwalb2009object} framework. In the deep learning era, some early detection systems \cite{yolo,redmon2017yolo9000,szegedy2014scalable,erhan2014scalable} also imply the high-level idea of \emph{multiple-instance-prediction}, while the methods are \textbf{not} proposal-based. For example, \emph{MultiBox} \cite{erhan2014scalable,szegedy2014scalable} directly predicts all the instances in an image patch; \emph{YOLO v1/v2} \cite{yolo,redmon2017yolo9000} generates multiple predictions for each cell (i.e. predicting all instances centered at a certain location). Special loss functions are also proposed in \cite{erhan2014scalable,szegedy2014scalable,yolo,redmon2017yolo9000} for set prediction, whose design purposes are similar to our \emph{EMD loss}. 

The most related previous work to us is \cite{stewart2016end} which introduces \emph{LSTM} to decode instance boxes in each grid of an image. Similar to our \emph{EMD loss}, they use \emph{Hungarian Loss} for multiple instance supervision. For post-processing, a box stitching method is employed to merge the predictions produced by adjacent grids. They mainly evaluated the method on head detection task, which shows some capability to predict crowded objects. However, since the method does not make use of proposals, it may have difficulty in detecting objects of various sizes/shapes, such as pedestrians or general objects. Moreover, the LSTM predictor is complex, which may be nontrivial to integrated in current state-of-the-art detection frameworks \cite{lin2017feature,lin2017focal,he2017mask} efficiently. 

\subsection{Network Architecture}
\label{sec:arch}
Theoretically, our approach can be applied to most of the state-of-the-art proposal-based detectors, no matter one-stage \cite{liu2016ssd,lin2017focal,yolov3} or two-stage \cite{ren2015faster,lin2017feature,he2017mask} frameworks.  In this paper, we choose \emph{FPN} \cite{lin2017feature} with \emph{RoIAlign} \cite{he2017mask} as a baseline detector to evaluate our method. In FPN, the \emph{Region Proposal Network (RPN)} branch takes the responsibility for proposal generation, while the \emph{RCNN} (or named \emph{RoI}) branch is used to predict the instance corresponding to the RoI proposal. So, our method is attached to the latter branch. From Sec.~\ref{sec:ourapproach}, easy to see that there is only one extra hyper-parameter in our approach -- $K$, the maximum cardinality of $\mathrm{G}(\cdot)$ (refer to Eq.~\ref{equ:pred}). In the rest of the paper, we let $K=2$, which we find is satisfied for almost all the images and proposals in many detection datasets like \emph{CrowdHuman} \cite{shao2018crowdhuman}, \emph{COCO} \cite{lin2014microsoft} and \emph{CityPersons} \cite{zhang2017citypersons}.

Fig.~\ref{fig:arch} (a) illustrates the usage of our method in FPN (only RCNN branch is shown). Based on the original architecture, only slight modifications need to be made: just attach an additional instance prediction head to the tail. \emph{EMD loss} is applied to the two predictions instead of the original loss. The \emph{refinement module} is optional; if applied, we use the refined results as the final predictions.

\section{Experiment}
\label{sec:exp}
In this section, we evaluate our approach from different perspectives. Intuitively, a detection algorithm specially optimized for crowded scenes tends to recall more instances, however, often have the risk of \textbf{increasing false positive predictions}. Our benchmarks focus on both of the opposite aspects. 

\paragraph{Datasets.}
An ideal object detector for crowded scenes should be robust to instance distributions, i.e. not only effective for crowded detections, but also stable to detect single/less-crowded objects. We adopt three datasets -- \emph{CrowdHuman} \cite{shao2018crowdhuman}, \emph{CityPersons} \cite{zhang2017citypersons} and \emph{COCO} \cite{lin2014microsoft} -- for comprehensive evaluations on heavily, moderately and slightly overlapped situations respectively. Table~\ref{tbl:datasets} lists the ``instance density'' of each dataset.  Since our proposed approach mainly aims to improve crowded detections. So, we perform most of the comparisons and ablations on \emph{CrowdHuman}. Note that the experiment on uncrowded dataset like COCO is to verify whether our method does harm to isolated object detection, \textbf{not for significant performance improvements}. 
\begin{table}[t]
   \centering
   \begin{tabular}{l|c|c}
      \toprule
      Dataset & \# objects/img & \# overlaps/img  \\
      \hline
      CrowdHuman \cite{shao2018crowdhuman} & 22.64  & 2.40 \\
      CityPersons \cite{zhang2017citypersons} & 6.47  & 0.32 \\
      COCO$^*$ \cite{lin2014microsoft} & 9.34  & 0.015 \\
      \bottomrule
   \end{tabular}
   \caption{\emph{Instance density} of each dataset.  The threshold for overlap statistics is $\mathrm{IoU} > 0.5$.  *Averaged by the number of classes.}
   \label{tbl:datasets}
\end{table} 
 
\paragraph{Evaluation metrics} We mainly take the following three criteria for different purposes:
\begin{itemize}
	\item \emph{Averaged Precision (AP)}, which is the most popular metric for detection. AP reflects both the precision and recall ratios of the detection results. In our experiment, we empirically find AP is more sensitive to the recall scores, especially on crowded dataset like \emph{CrowdHuman}. Larger AP indicates better performance.
	
	\item  \emph{$\text{MR}^{-2}$} \cite{dollar2012pedestrian}, which is short for \emph{log-average Miss Rate on False Positive Per Image (FPPI)} in $[10^{-2}, 100]$, is commonly used in pedestrian detection. $\text{MR}^{-2}$ is very sensitive to false positives (FPs), especially FPs with high confidences will significantly harm the $\text{MR}^{-2}$ ratio. Smaller $\text{MR}^{-2}$ indicates better performance. 
	
	\item \emph{Jaccard Index (JI)} \cite{liu2016ssd} is mainly used to evaluate the counting ability of a detector. Different from AP and $\text{MR}^{-2}$ which are defined on the prediction \emph{sequence} with decreasing confidences, JI evaluates how much the prediction \emph{set} overlaps the ground truths. Usually, the prediction set can be generated by introducing a confidence score threshold. In this paper, for each evaluation entry, we report the best JI score by exploring all possible confidence thresholds. We use the official SDK of \emph{CrowdHuman} \cite{shao2018crowdhuman} for JI calculation. Larger JI indicates better performance.
	
\end{itemize}

\paragraph{Detailed Settings. } 
\label{sec:detailed_setttings}
Unless otherwise specified, we use standard \emph{ResNet-50} \cite{he2016deep} pre-trained on \emph{ImageNet} \cite{russakovsky2015imagenet} as the backbone network for all the experiments. The baseline detection framework is \emph{FPN} \cite{lin2017feature}, while using \emph{RoIAlign} \cite{he2017mask} instead of original \emph{RoIPooling}. As for anchor settings, we use the same anchor scales as \cite{lin2017feature}, while the aspect ratios are set to $H:W = \{1:1, 2:1, 3:1\}$ for \emph{CrowdHuman} and \emph{CityPersons}, and $\{2:1, 1:1, 1:2\}$ for \emph{COCO}. For training, we use the same protocol as in \cite{lin2017feature}. The batch size is 16, split to 8 GPUs. Each training runs for 30 epochs. During training, the sampling ratio of positive to negative proposals for RoI branch is $1:1$ for CrowdHuman and $1:3$ for CityPersons and COCO. Multi-scale training and test are not applied; instead, the short edge of each image is resized to 800 pixels for both training and test.  All box overlap IoU thresholds (e.g. $\theta$ in Eq.~\ref{equ:gt_set}, NMS thresholds, and those in calculating evaluation metrics) are set to 0.5 by default. For our method, we use $K=2$ (see Eq.~\ref{equ:pred}). The \emph{refinement module} in Fig.~\ref{fig:arch} is enabled by default.

\subsection{Experiment on CrowdHuman}

\emph{CrowdHuman} \cite{shao2018crowdhuman} contains 15,000, 4,370 and 5,000 images for training, validation and test respectively. For fair comparison, we retrain most of the involved models with our own implementation under the same settings. Results are mainly evaluated on the validation set, using full-body benchmark in \cite{shao2018crowdhuman}. 

\paragraph{Main results and ablation study.} Table~\ref{tbl:crowd_ablation} shows the ablation experiments of the proposed methods in Sec.~\ref{sec:ourapproach}, including \emph{multiple instance prediction} with \emph{EMD loss}, \emph{set NMS} and \emph{refinement module}. The baseline is \emph{FPN} \cite{lin2017feature} using NMS (IoU threshold is 0.5) for post-processing. It is clear that our methods consistently improve the performances in all criteria. Especially, even without \emph{refinement module} our method still obtains \textbf{4.5\%} improvements in AP and \textbf{2.2\%} in JI, suggesting that more instances may correctly detected; more importantly, we find the $\text{MR}^{-2}$ ratio also improves, indicating that our model does not introduce more false predictions. The refinement module affects little on AP and JI, while further boosting  $\text{MR}^{-2}$ by $\sim$0.8\%, suggesting that the module mainly reduces false positives as we expected. 

\begin{table}[ht]
	\centering
	\begin{tabular}{ccc|ccc}
		\toprule
		  MIP & Set NMS & RM & AP/\% & $\text{MR}^{-2}$/\%  & JI/\% \\
		\hline
		 & & & 85.8 & 42.9 & 79.8 \\
		\hline 
		\checkmark & & & 87.4 & 42.8 &  80.8  \\
		\checkmark & \checkmark &  & 90.3 & 42.2 &  82.0  \\
		\checkmark & \checkmark & \checkmark & \textbf{90.7} & \textbf{41.4} &  \textbf{82.3}  \\
		\bottomrule
	\end{tabular}
	\caption{ Ablation experiments evaluated on \emph{CrowdHuman} validation set. The baseline model (the first line) is our reimplemented \emph{FPN} \cite{lin2017feature} with \emph{ResNet-50} \cite{he2016deep} backbone. \emph{MIP} -- multiple instance prediction with \emph{EMD loss}. \emph{RM} -- refinement module.  }
	\label{tbl:crowd_ablation}
\end{table}

\paragraph{Comparisons with various NMS strategies. }
In Fig.~\ref{fig:illustration}, since some instances are mistakenly suppressed by \emph{NMS}, one possible hypothesis is that the predictions may be improved by using different NMS strategies. Table~\ref{tbl:crowdhuman_nms} explores some variants. For na\"ive NMS, compared with the default setting (0.5), slightly enlarging the IoU threshold (from 0.5 to 0.6) may help to recall more instances, so AP increases; however, the $\text{MR}^{-2}$ index becomes much worse (from 42\% to 45.4\%), indicating that more false positives are introduced. \emph{Soft-NMS} \cite{softnms} can boost AP score, but no improvements are obtained in $\text{MR}^{-2}$ and JI. In contrast, our method achieves the best scores in all the three metrics even without \emph{refinement module}. 

\begin{table}[t]
	\centering
	\begin{tabular}{l|c|ccc}
		% \hline
		% \hline
		\toprule
		Method & IoU$^*$ & AP/\% & $\text{MR}^{-2}$/\%  & JI/\% \\
		\hline
		 & 0.3 & 72.3 & 48.5 & 69.6 \\
		 & 0.4 & 80.7 & 44.6 & 76.3 \\
		NMS & 0.5 & 85.8 & \underline{42.9} & \underline{79.8} \\
		 & 0.6 & \underline{88.1} & 45.4 & 79.4 \\
		 & 0.7 & 87.1 & 56.5 & 74.4 \\
		 & 0.8 & 82.8 & 68.5 & 62.6 \\
		\hline
		 Soft-NMS ~\cite{softnms} & 0.5 & 88.2 & 42.9 & 79.8 \\
		\hline
		Ours (w/o RM) & 0.5 & 90.3 & 42.2 &  82.0  \\
		Ours (with RM) & 0.5 & \textbf{90.7} & \textbf{41.4} & \textbf{82.3} \\
		\bottomrule
	\end{tabular}
	\caption{ Comparisons of different NMS strategies on \emph{CrowdHuman} validation set. The baseline model is \emph{FPN} \cite{lin2017feature}. $^*$IoU threshold for post-processing.  \emph{RM} -- refinement module. }
	\label{tbl:crowdhuman_nms}
\end{table} 

\paragraph{Comparisons with previous works. }
To our knowledge, very few previous works on crowded detection  report their results on \emph{CrowdHuman}. To compare, we benchmark two methods -- \emph{GossipNet} \cite{learningnms} and \emph{RelationNet} \cite{hu2017relation}  -- which are representative works categorized into \emph{advanced NMS} and \emph{re-scoring} approaches respectively (see Sec.~\ref{sec:limitations} for the analyses). For \emph{GossipNet}, we use the open-source implementation to benchmark\footnote{https://github.com/hosang/gossipnet}. And for \emph{RelationNet}, we re-implement the \emph{re-scoring} version\footnote{We use the \emph{re-scoring} version rather than \emph{NMS}, as NMS is clearly not suitable for crowded detection. We have checked \emph{COCO} \cite{lin2014microsoft} scores to ensure correct re-implementation.  }. All methods use \emph{FPN} \cite{lin2017feature} as the base detector with the same training settings. 

Table~\ref{tbl:crowdhuman_comp} lists the comparison results. Surprisingly, both \emph{RelationNet} and \emph{GossipNet} suffer from significant drop in AP and $\text{MR}^{-2}$. Further analyses indicate that the two methods have better recall ratio than baseline NMS for crowded objects (see Table~\ref{tbl:crowdhuman_recall}), however, tend to introduce too many false positive predictions. Though it is still too early to claim \cite{hu2017relation,learningnms} do not work on \emph{CrowdHuman} (we have not fully explored the hyper-parameters), at least the two methods are nontrivial for tuning. In contrast, our method is not only effective, but also very simple, as it has almost no additional hyper-parameters. 

Table~\ref{tbl:crowdhuman_comp} also compares a recent work \emph{AdaptiveNMS} \cite{adaptiveNMS}, which is an enhanced NMS strategy for crowded detection. In \cite{adaptiveNMS}, \emph{CrowdHuman} results based on \emph{FPN} are reported. Note that since the baseline are not aligned, we cannot make the direct comparison with our results. From the numbers, we find that our method can achieve significant improvement from a stronger baseline (especially in AP), in addition, the pipeline is much simpler. 

Table~\ref{tbl:crowdhuman_comp} also evaluate our method on the Cascade R-CNN \cite{cai2019cascadercnn} framework. 
We add the EMD loss and Set NMS into the last stage of Cascade R-CNN.
The results show our method can still boost the performance of Cascade R-CNN significantly on crowded datasets like CrowdHuman.
\begin{table}[ht]
   \centering
   \begin{tabular}{p{44mm}|p{6mm}<{\centering}p{10mm}<{\centering}p{6mm}<{\centering}}
   \toprule
       Method & AP/\% & $\text{MR}^{-2}$/\%  & JI/\% \\
       \hline
      FPN baseline & 85.8 & 42.9 & 79.8 \\
      FPN + Soft-NMS \cite{softnms} & 88.2 & 42.9 & 79.8 \\
       \hline
       RelationNet ~\cite{hu2017relation} (our impl.) & 81.6 & 48.2 & 74.6 \\
       GossipNet~\cite{learningnms}  (our impl.) &  80.4 & 49.4 & 81.6  \\
       \hline
       \textbf{Ours} & \textbf{90.7} & \textbf{41.4} & \textbf{82.3} \\
       \hline
       \hline
       FPN baseline (impl. by \cite{adaptiveNMS}) & 83.1 & 52.4 & --- \\
       AdaptiveNMS (impl. by \cite{adaptiveNMS}) & 84.7 & 49.7 & --- \\
       \hline
       \hline
       CascadeR-CNN \cite{cai2019cascadercnn} (our impl.) & 86.2 & 40.2 & 80.4 \\
       CascadeR-CNN + Ours & \textbf{90.6} & \textbf{38.7} & \textbf{83.9} \\
       \bottomrule
   \end{tabular}
   \caption{Comparisons of various crowded detection methods on \emph{CrowdHuman} validation set. All methods are based on \emph{FPN} detector \cite{lin2017feature}. Higher values of AP and JI indicate better performance, which is in contrast to the $\text{MR}^{-2}$.
      We use only one more stage in both of our Cascade R-CNN implementation instead of two for better performance in CrowdHuman.}
   \label{tbl:crowdhuman_comp}
\end{table}

\paragraph{Analysis on recalls.}
To further understand the effectiveness of our method on crowded objects, we compare the recalls of different approaches for both crowded and uncrowded instances respectively. Results are shown in Table.~\ref{tbl:crowdhuman_recall}. Note that recall relates to the confidence score threshold. For fair comparison, we use the thresholds corresponding to the best JI index for each entry respectively. From the table we find that for \emph{FPN} baseline/\emph{Soft-NMS}, recall of crowded objects is much lower than that of uncrowded objects, implying the difficulty of crowded detection. In contrast, our method greatly improves the recall ratio of the crowded instances (from 54.4\% to 63.3\%, by \textbf{8.9\%}), in addition, uncrowded recall is also slightly improved. 

\begin{table}[ht]
   \centering
   \begin{tabular}{p{24mm}|p{7mm}<{\centering}|p{10mm}<{\centering}p{10mm}<{\centering}p{10mm}<{\centering}}
   \toprule
   Method & Conf. & Total & Sparse  & Crowd \\
   \hline
   Ground truth & - & 99481 & 78665 & 20816 \\
   \hline
   FPN + & 0.7 & 80385   & 66880 & 13505 \\
   NMS  	& & (80.8\%) & (85.0\%) & (64.9\%) \\
   % NMS  	& & (80.8\%)\footnote{} & (85.0\%) & (64.9\%) \\
   \hline
   FPN + & 0.7 & 80385   & 66880 & 13505 \\
   Soft-NMS \cite{softnms} & & (80.8\%) & (85.0\%) & (64.9\%) \\
   \hline
   FPN + & 0.0 & 80196 & 65083 & 15113 \\
   GossipNet \cite{learningnms} & & (80.6\%) & (82.7\%) & (72.6\%) \\
   \hline
   FPN + & 0.5 & 75178 & 60083 & 15095 \\
   RelationNet \cite{hu2017relation} & & (75.6\%) & (76.4\%) & (72.5\%) \\
   \hline
   \textbf{Ours}  & 0.7 &  \textbf{83246} & \textbf{67716} & \textbf{15530} \\
            & & \textbf{(83.7\%)} & \textbf{(86.1\%)} & \textbf{(74.6\%)} \\
   \bottomrule
   \end{tabular}
   \caption{Detection boxes recalled on \emph{CrowdHuman} validation set. 
   Only boxes with confidence score higher than a certain threshold are taken into account.
   The confidence thresholds are subject to the best JI scores respectively and noted in the ``Conf'' column.
   Numbers in the last three columns indicate the number of recalled boxes.
   ``Crowd'' means the corresponding ground-truth box overlaps with some other 
   ground truth with IoU$>$0.5, otherwise marked ``Sparse''. 
   Note that recalls of \emph{Soft-NMS} \cite{softnms} are the same with the 
   NMS baseline, which is because the confidence threshold is relativity high 
   (0.7), thus NMS is roughly equivalent to Soft-NMS.  }

   \label{tbl:crowdhuman_recall}
\end{table}
% \footnotetext[3]{Please note that if recall $<$ JI, it's not a bug, because in \emph{CrowdHuman} SDK \cite{shao2018crowdhuman} JI is evaluated by averaging the JI score of each image, while in recall calculation the divisor is the number of ground truths of the whole dataset. }

\begin{figure*}
	\begin{center}
		\includegraphics[width=1.\linewidth]{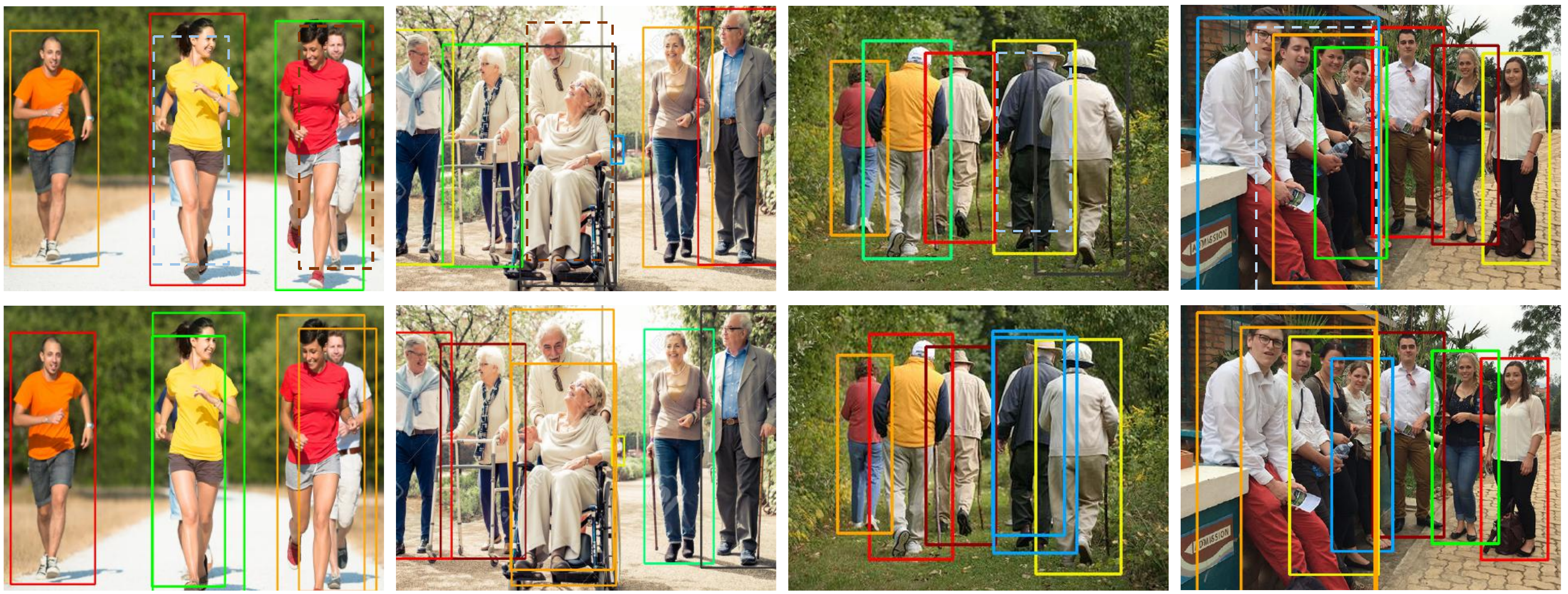}
	\end{center}
	\caption{Visual comparison of the baseline and our approach. The first row are the results produces by FPN with NMS. The last row are the results of our approach. The scores threshold for visualization is 0.3. Boxes with the same color stem from the identical proposal. The dashed boxes are the missed detection ones.}
	\label{fig:presentation}
\end{figure*}

\subsection{Experiments on CityPersons}

CityPersons~\cite{zhang2017citypersons} is one of the widely used benchmarks for pedestrian detection. The dataset contains 5, 000 images (2, 975 for training, 500 for validation, and 1, 525 for testing, respectively). Each image is with a size of 1024 $\times$ 2048. Following the common practice in previous works, all of the object detectors are trained on the training (\emph{reasonable}) subset and tested on the validation (\emph{reasonable}) subset with the enlarged resolution by 1.3$\times$ compared to the original one, which is slightly different from the settings we used for \emph{CrowdHuman} \cite{shao2018crowdhuman} and \emph{COCO} \cite{lin2014microsoft}. To obtain a better baseline, we follow the strategy proposed in ~\cite{chi2019pedhunter}, namely evolving ground-truths into proposals by jittering.Other hyper-parameters remains the same as that in Sec.~\ref{sec:detailed_setttings}.  

\begin{table}[ht]
\centering
\begin{tabular}{p{23mm}|p{9mm}<{\centering}|p{10mm}<{\centering}p{10mm}<{\centering}p{10mm}<{\centering}}
\toprule
 Method & Conf. & Total & Sparse  & Crowd \\
\hline
Ground truth & - & 1579 & 1471 & 108 \\
\hline
FPN & 0.6 & 1430   & 1366 & 64 \\
+NMS & & (90.6\%) & (92.9\%) & (59.3\%) \\
\hline
FPN  & 0.6 & 1430   & 1366 & 64 \\
 + Soft-NMS \cite{softnms} & & (90.6\%) & (92.9\%) & (59.3\%) \\
\hline

\textbf{Ours}  & 0.6 &  \textbf{1476} & \textbf{1380} & \textbf{96} \\
       	& & \textbf{(93.5\%)} & \textbf{(93.8\%)} & \textbf{(88.9\%)} \\
\bottomrule
\end{tabular}
\caption{Detection recalls on \emph{CityPersons} validation set. Please refer to the caption in Table~\ref{tbl:crowdhuman_recall} for the descriptions. Note that recalls of \emph{Soft-NMS} \cite{softnms} are the same as the NMS baseline, which is ascribed to the confidence threshold is relativity high (0.6), thus NMS is roughly equivalent to Soft-NMS. }

\label{tbl:citypersons_recall}
\end{table}

\paragraph{Qualitative results. }
Table~\ref{tbl:citypersons_eval} compares our method with \emph{FPN} baselines with na\"ive NMS and \emph{Soft-NMS} respectively. Our approach improves AP and $\text{MR}^{-2}$ by 0.9\% and 1.0\% respectively over the NMS baseline, indicating the effectiveness our method. Table~\ref{tbl:citypersons_eval} also lists some other state-of-the-art results on \emph{CityPersons}. Though it may be unfair for direct comparisons due to different hyper-parameter settings, however, at least it implies our method achieves significant gains over a relatively strong baseline. 

Table~\ref{tbl:citypersons_recall} further analyzes the recalls of different methods. Similar to those in \emph{CrowdHuman} (refer to Table~\ref{tbl:citypersons_recall}), our method mainly significantly boosts the recall on crowded objects -- from 64 increased to \textbf{96} out of a total of 108 instances in the validation set. The comparison further indicates our approach is very effective to deal with crowded scenes again. 
 
\begin{table}[!htbp]
   \centering
   \begin{tabular}{p{35mm}|p{13mm}<{\centering}|p{10mm}<{\centering}p{8mm}<{\centering}}
   % \begin{tabular}{l|c|cccc}
   % \setlength{\tabcolsep}{5mm}
   \toprule
   Method & Backbone & $\text{MR}^{-2}$ & AP \\
   \hline
   FPN + NMS & Res-50 & 11.7\% & 95.2\%  \\
   FPN + Soft-NMS \cite{softnms} &  & 11.8\% & 95.3\%  \\
   \hline
   \textbf{Ours}& Res-50 & \textbf{10.7\%} & \textbf{96.1\%}   \\
   \hline\hline
   AF-RCNN~\cite{zhang2017citypersons} &  & 12.8\% &---\\
   OR-CNN~\cite{zhang2018occlusion} & VGG-16 &11.0\% & ---\\
   Adaptive-NMS ~\cite{adaptiveNMS}  & &  10.8\% & ---  \\
   \hline
   FRCNN (our impl.) \cite{ren2015faster}  & Res-50 & 11.6\% & 95.0\% \\
   Repulsion Loss~\cite{repulseloss}&  & 11.6\% & --- \\
   \bottomrule
   \end{tabular}
   \caption{Comparisons of different methods on \emph{CityPersons} validation set. All models are evaluated with enlarged resolution of 1.3$\times$ compared to the original size. Models in the upper part are trained by ours with the same \emph{FPN} base detector. Models in the lower part are trained with other protocols. }
   \label{tbl:citypersons_eval}
\end{table}

\subsection{Experiments on COCO}

According to Table~\ref{tbl:datasets}, the \emph{crowdedness} of \emph{COCO} \cite{lin2014microsoft} is very low, which is out of our design purpose. So, we \textbf{do not} expect a significant performance gain on COCO. Instead, the purpose of introducing COCO is to verify: 1) whether our method generalizes well to multi-class detection problems; 2) whether the proposed approach is robust to different crowdedness, especially to \emph{isolated} instances.  

Following the common practice of \cite{lin2017feature,lin2017focal}, we use a subset of 5000 images in the original validation set (named \emph{minival}) for validation, while using the remaining images in the original training and validation set for training. Table~\ref{tbl:mscoco_eval} shows the comparisons with FPN and FPN+\emph{Soft-NMS} baselines. Moderate improvements are obtained, e.g. 1.0\% better than na\"ive NMS and 0.5\% better than Soft-NMS in AP. Interestingly, large objects achieve the most significant improvement (see $\text{AP}_L$ in Table~\ref{tbl:mscoco_eval}), which may be because larger objects are more likely to overlap. The experiment suggests our method is not only very effective on crowded scenes, but also able to deal with multiple classes and isolated instances without performance drop. 

\begin{table}[t]
   \centering
   \begin{tabular}{p{15mm}|p{8mm}<{\centering}p{6mm}<{\centering}p{6mm}<{\centering}|p{6mm}<{\centering}p{6mm}<{\centering}p{6mm}<{\centering}}
   \toprule
   Method & AP & $\text{AP}_{50}$ & $\text{AP}_{75}$ &  $\text{AP}_{S}$ & $\text{AP}_{M}$  & $\text{AP}_{L}$\\
   \hline
   FPN & 37.5 & 59.6 & 40.4 & 23.0 & 41.2 & 48.6 \\
   Soft-NMS & 38.0 & 59.4 & 41.5 & 23.3 & 41.8 & 49.0 \\
   \hline
   \textbf{Ours} & \textbf{38.5} & 60.5 & 41.5 & 23.0  &  41.8  & 50.1 \\

   \bottomrule
   \end{tabular}
   \caption{Comparisons on \emph{COCO}~\cite{lin2014microsoft} \emph{minival} set. All models are based on \emph{FPN} detector. Results are evaluated in all the 80 classes.  }
   \label{tbl:mscoco_eval}
\end{table}

\section{Conclusion}
In this paper, we propose a very simple yet effective proposal-based object detector, specially designed for crowded instance detection. The method makes use of the concept of \emph{multiple instance prediction}, introducing new techniques such as \emph{EMD loss}, \emph{Set NMS} and \emph{refinement module}. Our approach is not only effective, but also flexible to cooperate with most state-of-the-art proposal-based detection frameworks; in addition, also generalizes well to less crowded scenarios. 

{\small
\bibliographystyle{ieee_fullname}
\bibliography{egbib}
}

\newpage
\section*{A. CrowdHuman Testing Benchmark}
The \emph{CrowdHuman}\cite{shao2018crowdhuman} testing subset has $5, 000$ images and the 
annotations of testing subset have not yet been released.
To push the upper-bound of object detection research,
the \emph{Detection In the Wild Challenge} was held in CVPR 2019.
The \emph{CrowdHuman} testing subset is served as a benchmark in this challenge and 
this allows us to compare our approach to state-of-the-art 
methods on \emph{CrowdHuman}.

To improve the performance of our approach, we replace the 
ResNet-50\cite{he2016deep} with a larger model: SEResNeXt101\cite{hu2018squeeze,xie2017aggregated}, 
the short edge of test images are resized to 1200 pixels
and all other settings are the same as described in our paper. 
We then submit our result to the test server and find that our method outperforms 
all the results in this challenge.
The results are shown in Table.\ref{tab:ledaerboard}, and the full leaderboard 
is accessible on the official website of CrowdHuman Track Leaderboard
\footnote{https://www.objects365.org/crowd\_human\_track.html}.

\begin{table}[htp]
   \begin{center}
   \begin{tabular}{c|l|l|p{8mm}<{\centering}}
   \hline\hline
   Rank &Team Name&Institution&mJI/\%\\
   \hline\hline
   1 & zack0704 & Tencent AI Lab & 77.46 \\
   2 & boke & Sun Yat-Sen University & 75.25 \\
   3 & ZNuanyang & Zhejiang University & 74.46 \\
   \hline\hline
   \multicolumn{2}{c|}{Method} & Backbone & mJI/\%\\
   \hline\hline
   \multicolumn{2}{c|}{Baseline} & ResNet-50 & 72.20 \\
   \hline
   \multicolumn{2}{c|}{Ours} & ResNet-50  & 76.60\\
   \multicolumn{2}{c|}{Ours} & SEResNeXt101 & {\bf 77.74}\\
   \hline
   \end{tabular}
   \end{center}
   \caption{Part of the leaderboard and our results. 
   The baseline model is our reimplemented FPN\cite{lin2017feature}.}
   \label{tab:ledaerboard}
\end{table}

\section*{B. Ablation on Number of Heads}
For completeness, we further explore the only hyper-parameter of our method: $K$
in this section.
In our paper, we let $K=2$ because we find it is satisfied for almost all the images
and proposals in \emph{CrowdHuman}.
If we make the $K$ larger, the network will be able to detect instances in more 
crowded scenes.
To explore the performance under the different values of the $K$, an experiment is 
conducted on the \emph{CrowdHuman} dataset and all the settings remain the same as 
described in our paper except the value of $K$ is changed.
We show the results of different $K$ values in the Table.\ref{tab:k3}.

\section*{C. More Results of Our Method}
In this section we will show more results of our method on a video from YouTube and 
the \emph{CrowdHuman} validation dataset.
The visualization thresh is set to 0.7 to remove the redundant boxes in the results.

The video is in the attached file, and the results on the \emph{CrowdHuman} 
validation dataset is shown in the Figure.\ref{fig:presentation}.

\makeatletter
\makeatother

\begin{table}
   \begin{center}
   \begin{tabular}{c|c|c|c}
   \hline\hline
   & AP/\% & $\text{MR}^{-2}$/\%  & JI/\% \\
   \hline\hline
   $K = 1$ & 85.8 & 42.9 & 79.8 \\
   $K = 2$ & {\bf 90.7} & {\bf 41.4} & {\bf 82.3 }\\
   $K = 3$ & {\bf 90.7} & 41.6 & 82.1 \\
   \hline
   \end{tabular}
   \end{center}
   \caption{
      Ablation experiments evaluated on the \emph{CrowdHuman} validation set.
      It worth noting that if K = 1, the architecture is the same as the
      \emph{single-instance-prediction} baseline.}
   \label{tab:k3}
\end{table}
\quad

\section*{D. Our Method in One-Stage Detector}
To verify the effectiveness of our method in one-stage detectors, 
we conducted experiments on RetinaNet\cite{lin2017focal}.
We adopt standard \emph{ResNet-50} pre-trained on \emph{ImageNet} as the backbone network.
The anchor ratios are set to $H:W = \{1:1, 2:1, 3:1\}$ because of the shape of
human instances.
We use a batch size of 16 pictures over 8 GPUs, and train the network with
a learning rate of 0.005 for 50 epochs.
The other hyperparameters are kept the same as \cite{lin2017focal}.
For our method, we predict two predictions based on each anchor.
And the Cross-Entropy Loss in EMD Loss is replaced by Focal Loss\cite{lin2017focal}.
The results are shown in Tabel\ref{tbl:retina},
which indicates that our method can also have gain in one-stage detectors.
\begin{table}[t]
   \centering
   \begin{tabular}{l|c|c|c}
   \toprule
   Method & AP/\% & $\text{MR}^{-2}$/\%  & JI/\% \\
   \hline
   RetinaNet Baseline & 82.0 & 56.3 & 73.2\\
   \hline
   \textbf{Ours} & \textbf{82.7} & \textbf{54.7} & \textbf{74.0}\\
   \bottomrule
   \end{tabular}
   \caption{
      Experiment on RetinaNet\cite{lin2017focal}.
      All hyperparameters between the RetinaNet baseline and our method
      remain the same.
      The results show that one-stage detectors can also benefit from our
      method.}
   \label{tbl:retina}
\end{table}

\newpage
\begin{figure*}
	\begin{center}
		\includegraphics[width=1.\linewidth]{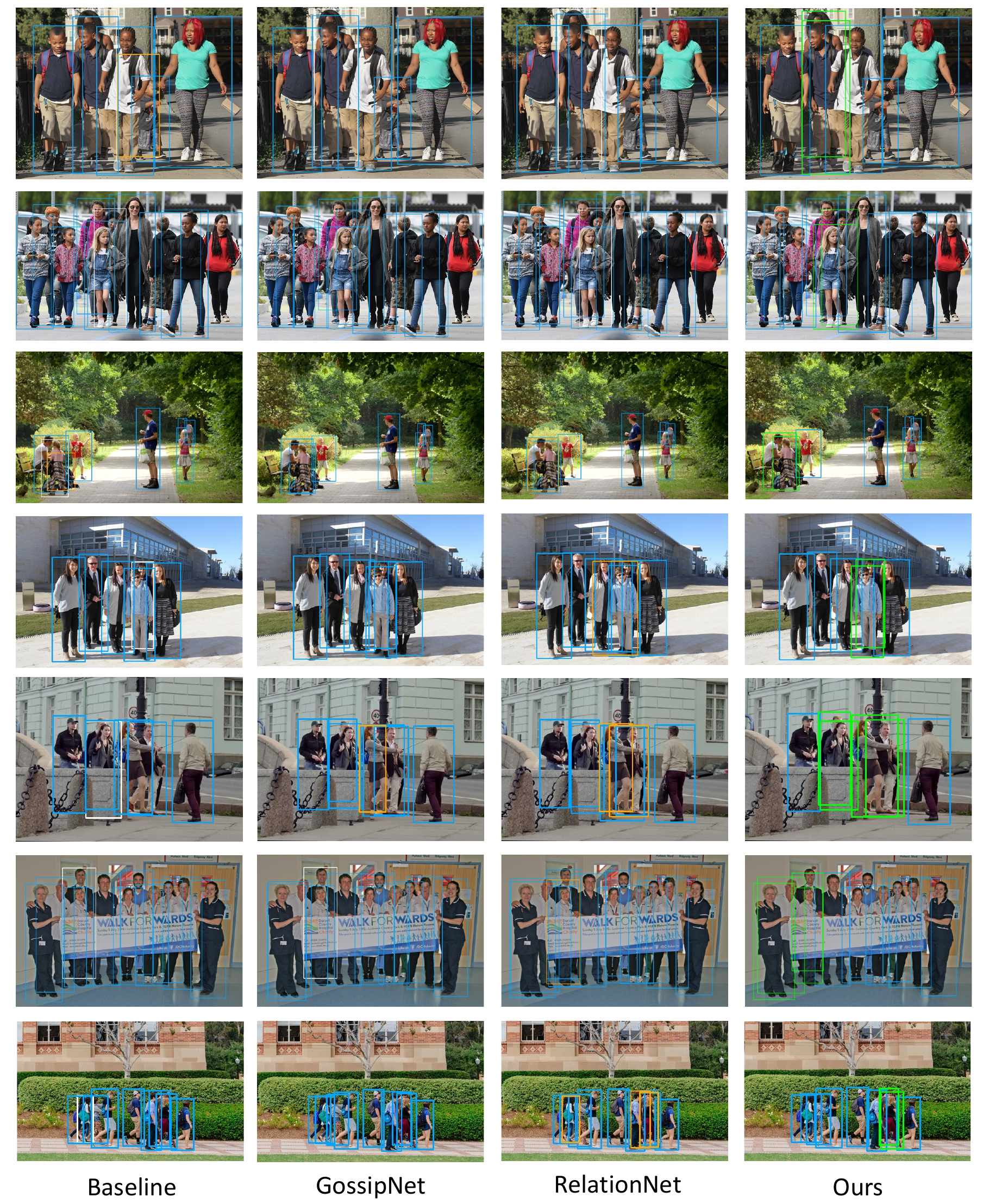}
	\end{center}
   \caption{Visual comparison of the baseline, GossipNet, RelationNet, and our 
   approach. 
   The blue boxes are the detection results, the white boxes are the missed 
   instances, and the orange boxes are redundant boxes. 
   The green boxes in our method are multiple predictions form one proposal.}
	\label{fig:presentation}
\end{figure*}

\end{document}